\documentclass[10pt,twocolumn,letterpaper]{article}
\usepackage{cvpr}

\usepackage{times}
\usepackage{epsfig}
\usepackage{graphicx}
\usepackage{amsmath}
\usepackage{amssymb}
\usepackage{arydshln}
\usepackage{mathtools}
\usepackage{multirow}
\usepackage{enumitem}
\usepackage{algorithm}
\usepackage{algpseudocode}
\usepackage{color}
\usepackage{wrapfig}

\newcommand{\keypoint}[1]{\vspace{0.1cm}\noindent\textbf{#1}\quad}
\newcommand{\cut}[1]{}
\DeclareMathAlphabet\mathbfcal{OMS}{cmsy}{b}{n}

\makeatother
\usepackage[pagebackref=true,breaklinks=true,colorlinks,bookmarks=false]{hyperref}

\cvprfinalcopy 

\def\cvprPaperID{5096} 

\ifcvprfinal\fi
\begin{document}

\title{StyleMeUp: Towards Style-Agnostic Sketch-Based Image Retrieval}
\author{Aneeshan Sain\textsuperscript{1,2} \hspace{.2cm} Ayan Kumar Bhunia\textsuperscript{1} \hspace{.2cm} Yongxin Yang\textsuperscript{1,2} \hspace{.2cm}  Tao Xiang\textsuperscript{1,2}\hspace{.2cm}  Yi-Zhe Song\textsuperscript{1,2} \\
\textsuperscript{1}SketchX, CVSSP, University of Surrey, United Kingdom. \\ \textsuperscript{2}iFlyTek-Surrey Joint Research Centre on Artificial Intelligence.\\
{\tt\small \{a.sain, a.bhunia, yongxin.yang, t.xiang, y.song\}@surrey.ac.uk}
}


\maketitle
\ifcvprfinal\thispagestyle{empty}\fi

\begin{abstract}

Sketch-based image retrieval (SBIR) is a cross-modal matching problem which is typically solved by learning a joint embedding space where the semantic content shared between photo and sketch modalities are preserved. However, a fundamental challenge in SBIR has been largely ignored so far, that is, sketches are drawn by humans and considerable style variations exist amongst different users. An effective SBIR model needs to explicitly account for this style diversity, crucially, to generalise to unseen user styles. To this end, a novel style-agnostic SBIR model is proposed. Different from existing models, a cross-modal variational autoencoder (VAE) is employed to explicitly disentangle each sketch into a semantic content part shared with the corresponding photo, and a style part unique to the sketcher. Importantly, to make our model dynamically adaptable to any unseen user styles, we propose to meta-train our cross-modal VAE by adding two style-adaptive components: a set of feature transformation layers to its encoder and a regulariser to the disentangled semantic content latent code. With this meta-learning framework, our model can not only disentangle the cross-modal shared semantic content for SBIR, but can adapt the disentanglement to any unseen user style as well, making the SBIR model truly style-agnostic. Extensive experiments show that our style-agnostic model yields state-of-the-art performance for both category-level and instance-level SBIR. 

\end{abstract}

 
\vspace{-0.4cm}
\section{Introduction}

Sketch as an input modality has been proven to be a worthy complement to text for photo image retrieval \cite{collomosse2019livesketch, dey2019doodle, sketch2vec}. Its precision in visual description is particularly useful for fine-grained retrieval, where the goal is to find a specific object instance rather than category \cite{sain2020cross, song2017deep, semi-fgsbir}. Research has flourished in recent years, where the main focus has been on addressing the sketch-photo domain gap ~\cite{umar2019goal, ha2017neural, sain2020cross} and data scarcity~\cite{bhunia2020sketch, semi-fgsbir, dey2019doodle, pang2019generalising, dutta2019semantically}. Thanks to these combined efforts, reported performances have already shown promise for practical adaptation. 

\begin{figure}[t]
\begin{center}
    \includegraphics[width=0.9\linewidth]{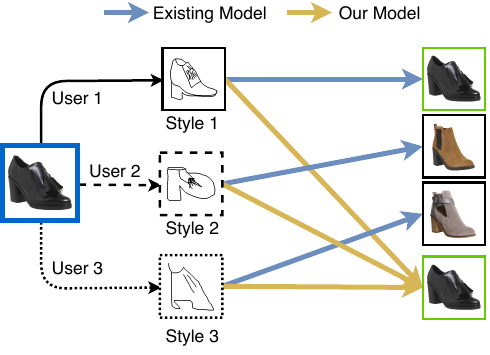}
    \vspace{-0.3cm}
    \caption{Owing to subjective interpretation, different users sketch the same object instance (a shoe here) very differently. Without considering this style diversity, an existing SBIR model   yields completely different results for different sketches. With our style-agnostic model, the same intended object is retrieved. }
    \vspace{-1.0cm}
\end{center}
\label{fig:Fig1}
\end{figure}

However, there is an important issue that has largely been ignored so far and has impeded the effectiveness of existing SBIR models -- sketches are drawn by humans and there exists considerable style variations amongst users (Fig.~\ref{fig:Fig1}). This is a result of subjective interpretation and different drawing skills of different users. Consequently, even with the same object instance as reference,  sketches of different users can look drastically different as shown in the example in Fig.~\ref{fig:Fig1}.  Existing SBIR models \cite{liu2017deep, collomosse2019livesketch,song2018learning, pang2019generalising, sain2020cross}  focus primarily on bridging the gap between the photo and sketch modalities. This is typically achieved by learning a joint embedding space where only the common semantic content part of a matching photo-sketch pair are preserved for matching. However, the large style variations of different users mean that the shared common semantic content can also vary (e.g., Fig.~\ref{fig:Fig1} shows that different users may choose to depict different characteristics of the same shoe). Crucially, it can vary in an unpredictable way -- a commercial SBIR model will be used mostly by  users whose sketches have never been used for model training. These models are thus poorly equipped to cope with this style diversity and unable to generalise to new user styles.      

In this paper, a novel style-agnostic SBIR framework is proposed which explicitly accounts for the style diversity and importantly can adapt dynamically to any unseen user styles without any model retraining. Different from existing SBIR models which focus solely on the shared semantic content between the photo and sketch modality and discard the modal-specific parts, we argue that in order to effectively deal with the style variations unique to the sketch modality, a disentanglement model is needed. With such a model, the user style can be modelled explicitly, making way for better generalisation.


The core of our style-agnostic SBIR framework is thus a disentanglement model, that takes a sketch or photo image as input and decomposes its content into a cross-modal shared semantic part to be used for retrieval,  and a modal-specific part -- in case of sketch, it corresponds to the user's drawing style.  Disentangling sketching styles is however a challenging task. Existing style disentanglement methods usually cater to problems where the style information carry less variance (e.g., schools of art, building styles, etc) and hence is comparatively easier to separate \cite{kotovenko2019content}. For sketches however, we are faced with much larger \textit{variability} where each user has a \textit{unique} style that can manifest itself in different ways for different object instances. The disentanglement should thus be able to dynamically adapt to new user styles and new object categories {for better generalisation}. To this end, we propose a novel disentanglement model with meta-learning, that\cut{ to make it} generalises to unseen user styles. 

Concretely, we employ a cross-modal {translation} variational autoencoder (VAE) framework \cite{kingma2013auto} to project a sketch/photo into its modal-invariant semantic part, and modal-specific part. The VAE is used for both sketch reconstruction and translation as well as sketch-to-photo translation to exploit the shared semantic content across both modalities and styles. To make the disentanglement dynamically adaptable and generalisable, a popular gradient-based meta-learning model namely model-agnostic meta-learning (MAML) \cite{finn2017MAML} is adopted. Designed for few-shot learning, the original  MAML cannot be directly applied. We thus introduce two new components as shown in Fig.~\ref{fig:Fig2}: (i) A set of feature transformation layers sitting between the VAE encoder layers for the encoder adaptation, and  (ii) a regulariser designed to adapt the disentangled modal-invariant semantic content part of the latent code produced by the encoder. 
Both component's parameters are meta-learned using MAML for fast adaptation to new style/categories sampled during episodic training of MAML. Once trained, these two components are responsible for adaptation to new/unseen user styles and object categories/instances, therefore achieving style-agnostic SBIR. 

Our contributions are as follows:
(a) For the first time, we propose the concept of style-agnostic SBIR to deal with a largely neglected user style diversity issue in SBIR.
(b) We introduce a novel style-agnostic SBIR framework based on disentangling a photo/sketch image into a  modal-invariant semantic content part suitable for SBIR and a model-specific part that needs to be explicitly modelled in order to minimise its detrimental effects on retrieval. (c) To make the disentanglement generalisable to unseen user styles and object categories/instances,  feature transformation layers and latent modal-invariant code regulariser are introduced to a VAE, both of which are meta-learned using a MAML framework for style/category/instance adaptation. 
(d) Extensive experiments show that state-of-the-art performances can be achieved as a direct result of the style-agnostic design.

\vspace{-0.2cm}
\section{Related Works}
\vspace{-0.1cm}
\noindent \textbf{Category-level SBIR:} 
Category-level SBIR tasks accept a sketch-query with an aim to retrieve photos of the same category \cite{sangkloy2016sketchy, collomosse2017sketching}. Early approaches deploy {handcrafted descriptors}~\cite{tolias2017asymmetric} such as  SIFT \cite{lowe1999object}, Gradient Field HOG \cite{hu2013performance}, Histogram of Edge Local Orientations \cite{saavedra2014sketch} or Learned Key Shapes \cite{saavedra2015sketch}, for constructing local \cite{hu2013performance} or global \cite{qi2015making} joint photo-sketch representations. Most recent approaches are based on deep learning~\cite{liu2017deep, collomosse2019livesketch}. They typically employ Siamese-like neural networks with ranking losses, like triplet loss~\cite{yu2016sketch} to learn a joint embedding space for both the sketch and photo modalities. Contemporary research directions also include zero-shot SBIR \cite{yelamarthi2018zero, dey2019doodle} where a model aims to generalise across disjoint training and testing classes \cite{dutta2019semantically}, alleviating annotation costs. Sketch-photo hashing \cite{liu2017deep, zhang2018generative}  on the other hand embeds to binary hash-codes instead of continuous vectors for computational ease.


\noindent \textbf{Fine-grained SBIR:} As opposed to category-level SBIR, fine-grained SBIR (FG-SBIR) \cite{song2018learning, pang2019generalising, sain2020cross} is directed towards instance-level sketch-photo matching. Starting with deformable-part models~\cite{li2014fine}, various deep approaches have surfaced with the advent of new FG-SBIR datasets \cite{song2017deep, yu2016sketch, huang2019swire}. Yu \etal \cite{yu2016sketch} introduced a deep triplet-ranking model that learnt a joint sketch-photo embedding space. This was further enhanced via attention based techniques with advanced higher order retrieval loss \cite{song2017deep}, hybrid generative-discriminative cross-domain image generation \cite{pang2017cross}, textual tags \cite{song2017fine} and employing mixed modal jigsaw solving for a better pre-training strategy \cite{pang2020solving}. While Sain \etal~\cite{sain2020cross} explored cross-modal hierarchical co-attention amongst sketch-photo regions, Bhunia \etal~\cite{bhunia2020sketch} employed reinforcement learning in an early retrieval scenario. 
These models focus on learning a joint embedding space where only the modal-invariant shared semantic content of a matching photo-sketch pair is preserved for both modalities. However, without explicitly modelling the modal-specific parts, particularly for sketches the user styles, these models cannot generalise well to unseen objects and user styles. 
\cut{In this paper, we work towards a novel training paradigm for FG-SBIR, where a photo-sketch representation is meta-learned by exploiting shared knowledge across tasks such that every representation is equipped with a consolidated knowledge of the target photo along with probabilistic variations of the query-sketch itself.}

\keypoint{Disentangled representation learning:}
Learning a disentangled representation would require modelling distinct informative factors in the variations of data~\cite{ding2020guided}. Starting from generic frameworks like combining auto-encoders with adversarial training \cite{mathieu2016disentangling}, this disentanglement paradigm has been successfully applied to recognition~\cite{wang2018orthogonal, peng2017reconstruction}, image-to-image translation~ \cite{yang2019disentangling} and image-editing \cite{jiang2019disentangled, wei2018real} tasks. While, InfoGAN~\cite{chen2016infogan} optimises mutual knowledge between latent variables, $\beta$-VAE~\cite{higgins2016beta} balances a hyperparameter $\beta$ to learn independent data generative factors for disentanglement in an unsupervised setting. These methods however lack interpretability, with the relevance of each learned factor being uncontrollable. A few recent works include joint disentanglement and adaptation module trained in a cross-domain cycle-consistency paradigm \cite{zou2020joint} or multi-scale spatial-temporal maps with a cross-verified disentangling strategy \cite{niu2020video} or   adversarial parameter estimation \cite{pizzati2020model}. 
None of such methods however has worked towards disentangling features for sketches to assist in SBIR. Furthermore, none has the ability to adapt the disentanglement  dynamically for new user styles and object instances, which our meta-learning based cross-modal disentanglement VAE is designed for. 

\keypoint{Meta-Learning:}
Meta-learning aims to acquire transferable knowledge from different sample training-tasks to help adapt to unseen tasks with only a few training samples \cite{tim2020metaSurvey}. Most existing meta-learning methods \cite{vinyals2016matching,snell2017prototypical,chen2019closerFew} are designed for few-shot image classification and thus are suitable for our problem of meta-learning of a generalisable cross-modal disentanglement model. 
The popular gradient/optimisation based meta-learning method MAML~\cite{finn2017MAML}, however is general enough to be adapted to our problem. MAML trains a base model on a set of source tasks to learn good initialisation parameters that adapts quickly to new tasks during training, over just a few gradient descent updates. Since its first introduction, various modifications have been proposed including Meta-SGD \cite{li2017metaSGD}, MAML++~\cite{antoniou2018trainMAML}, latent embedding optimization (LEO)~\cite{rusu2019LEO} and  uncertainty-induced MAML for continual learning~\cite{finn2019online}. Among them, those designed for domain adaptation~\cite{tseng2020cross} or domain generalization \cite{li2017learning,balaji2018metareg}
are the most relevant to our work. Different from them, our model uniquely addresses the cross-modal SBIR problem via meta-learning a disentanglement VAE for \textit{generalising} better onto unseen user styles and object instances. 


\vspace{-0.2cm}
\section{Methodology}
\vspace{-0.2cm}
\keypoint{Overview:} 
We aim to devise a SBIR framework that learns to model the diversity in sketching-styles corresponding to the same object category (for category-level SBIR~\cite{collomosse2017sketching}) or instance (for FG-SBIR~\cite{sain2020cross}). To this end, we design a style agnostic disentanglement model which decomposes the content of a photo/sketch image into a modal-invariant semantic part suitable for cross-modal matching, and a modal-specific part which is a distractor to SBIR but needs to be modelled explicitly to assist in the disentanglement. The disentanglement model is a cross-modal VAE that learns to embed a photo/sketch image to reconstruct them either in its original modality or to the other modality. 

\begin{figure}[t]
\begin{center}
  \includegraphics[width=\linewidth]{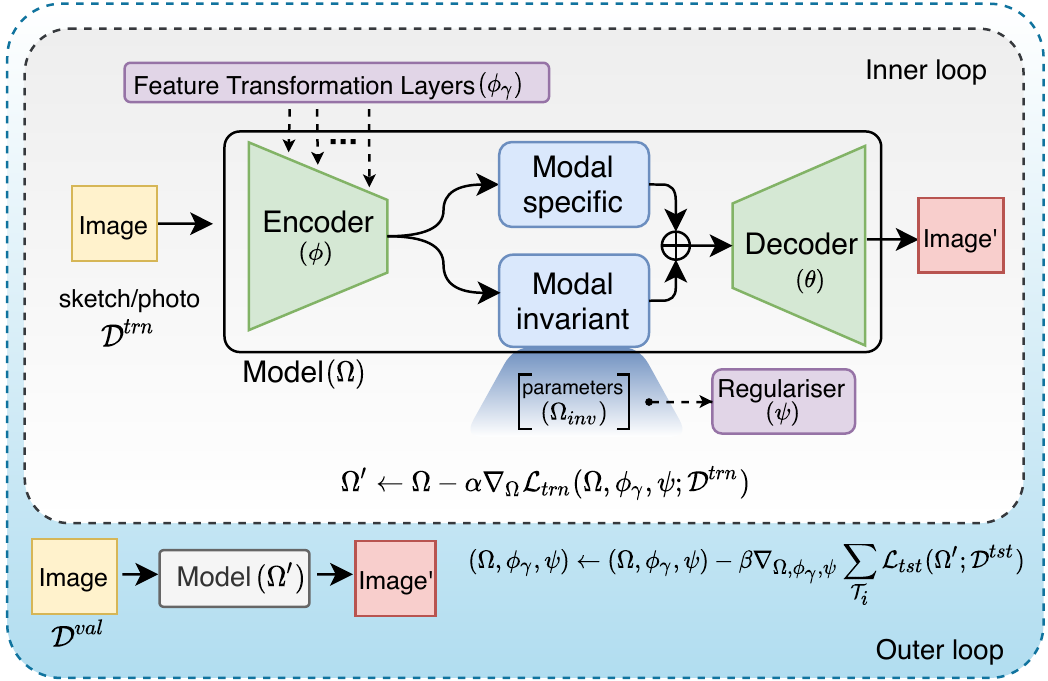}
\end{center}
  \vspace{-0.3cm}
  \caption{Our core model is a VAE framework that disentangles the modal variant and invariant semantics in a sketch in a cross-modal translation setting. While a regulariser network regularises parameters of the invariant component ($\Omega_{inv}$), feature transformation (FT) layers aid in style-agnostic encoding following a meta-learning paradigm. 
  }
\vspace{-0.6cm}
\label{fig:Fig2}
\end{figure}

Formally,  we are given  a set of  $C = \{C_1,C_2,\cdots,C_M\}$ categories ($M\geq1$) where every category $C_i$ has 
$d^i = \{d^i_1,d^i_2,\cdots,d^i_{N_i}\}$ ($N_i \geq1$) data-points. Every data-point $d^i_j |_{j=1}^{N_i}$ corresponds to a sketch($s$)-photo($p$) pair i.e. $\{s^i_j, p^i_j\}$. For FG-SBIR, every photo instance is treated as a category with multiple sketch-styles paired with the same photo ($p^i$), i.e $\forall d^i_j, j \in [1,N^i] \; p_j = p^i$. 
Feeding each data point to the encoder of the VAE, a latent code is obtained for both the photo and sketch. Our model (Fig.~\ref{fig:Fig2}) aims to  disentangle the latent code  into modal-invariant and modal-specific  components. The former corresponds to the semantic content of the object and thus should be used for cross-modal matching. It is subjected to a triplet loss so as to  minimise the distance of from a sketch sample ($s$) to its matching photo sample ($p+$), while increasing that to an unmatched one ($p-$). 
Such a model is trained in a meta-learning framework for better generalisation. Once trained, during inference it uses the learnt encoder and the modal-invariant component to produce a style-agnostic embedding function $ \mathcal{F(\cdot)}: \mathbb{R}^{H \times W \times 3} \rightarrow \mathbb{R}^{d_h}$ to map a rasterised sketch/photo having height $H$ and width $W$ to a $\mathbb{R}^{d_h}$ feature for matching.

\vspace{-0.1cm}
\subsection{Disentanglement by Cross-modal Translation}
\vspace{-0.1cm}
\label{sec:cmd}
Our disentanglement model is built upon a VAE framework~\cite{kingma2013auto} for both intra-modal reconstruction and cross-modal translation. The original VAE model produces a latent representation by optimising the variational lower bound on log-likelihood of the data 
\vspace{-0.1cm}
\begin{equation}
\label{equ:mainVAE}
    \operatorname{log} \mathbf{p}(x) \geq \mathbb{E}_{z\sim \mathbf{q}(z|x)}[\operatorname{log}\mathbf{p}(x|z) ] - D_{\text{KL}}(\mathbf{q}(z|x)||p(z)),
\vspace{-0.1cm}
\end{equation}
where  $D_{\text{KL}}(\cdot,\cdot)$ is the Kullback-Leibler (KL) divergence, and the conditional probability distributions $\mathbf{q}(z|x), \mathbf{p}(x|z)$ refer to the encoder and decoder respectively, both parameterised by neural networks. The distribution $\mathbf{p}(z)$ is the prior on the latent space, modeled as $\mathcal{N}(z|\mathbf{0}, \mathbf{I})$. The encoder returns mean $\mu$ and variance $\sigma^2$ of a normal distribution, such that $z \sim \mathcal{N}(\mu, \sigma^2)$.
Unlike this formulation which takes a single distribution into account, cross-modal training requires considering at least two modalities. Following \cite{spurr2018cross} we extend Eq.~\ref{equ:mainVAE} to the multi-modality case as:
\vspace{-0.2cm}
\begin{equation}
    \operatorname{log} \mathbf{p}(x_o) \geq \mathbb{E}_{z\sim \mathbf{q}(z|x_i)}[\operatorname{log}\mathbf{p}(x_o|z) ] - D_{\text{KL}}(\mathbf{q}(z|x_i)||p(z)),
\vspace{-0.1cm}
\end{equation}
where $x_i, x_o$ represents samples from input and output modalities respectively. It shows that the input and output samples can be decoupled via a joint embedding space shared by both sketch and photo modalities. This shared space thus allows for both same as well as cross-modality translations. Cross-modal training importantly creates a manifold which places objects to a high-dimensional space enriched with sketch-photo semantic relevance.

So far, we have described a VAE for cross-modal as well as intra-modal image translation. Our objective is however to disentangle the image content into modal-invariant and model-specific parts. Such a disentanglement takes place in the latent space produced by the encoder. More specifically, our CNN encoder $Enc_{\phi}$ projects an input $I$ into two parts: a modal-invariant  component ($z_{inv}$), and parameters mean ($\mu$) and variance ($\sigma$) for a variable (modal-specific) component ($z_{var}$). Essentially the variable component is modelled via an independent unit Gaussian distribution as  $z_{var} = \mu + \sigma \odot \mathcal{N}(0,1)$ where $z \in \mathbb{R}^{d_h}$. Combining both components we thus obtain our final latent variable $z_f = z_{var} \oplus z_{inv}$, where $\oplus$ represents element-wise summation. $z_f$ is then fed to the decoder for reconstruction as $\hat{I} = Dec_\theta(z_f)$. 
Such a VAE model is trained by optimising the sum of reconstruction ( $\mathcal{L}_{\text{rec}}$) and KL divergence ($\mathcal{L}_{\text{KL}}$) losses via gradient descent, with:
\vspace{-0.2cm}
\begin{equation}
\label{equ: vae}
\begin{aligned}
    &\mathcal{L}_{\text{rec}}(\phi,\theta) = -\mathbb{E}_{\mathbf{q}(z_f|I)}[\log \; \mathbf{p}(I|z_f)],  \\
    &\mathcal{L}_{\text{KL}} = D_{\text{KL}}[\mathbf{q}_\phi(z_f|I) \;||\; \mathbf{p}(z_f)],
\end{aligned}
\vspace{-0.1cm}
\end{equation}
where the prior over latent variables is a centered isotropic multivariate Gaussian, $\mathbf{p}_\theta(z_f) = \mathcal{N}(z_f;0,I)$.
In practice however, we simplify $\mathcal{L}_{\text{rec}} = \|\hat{I} - I\|_2$.

Besides sketch-photo translation we perform cross-style translation between two sketches of the same object, which ensures modelling the style diversity.
Concretely, given two sketches $s_j$ and $s_k$ we model the latent feature of $s_j$ as $z_f^{s_j*} = z_{inv}^{s_j} \oplus z_{var}^{s_k}$, and then we reconstruct it by $\widehat{s_j}^* = Dec_\theta(z_f^{s_j*})$, where $s_k$ is another randomly chosen style of the same object ($j,k \in [1,N_i] \;; j\neq k$). 
Accordingly we obtain the sum of all reconstruction losses as $\mathcal{L}_{\text{rec}}$ and sum their corresponding KL-divergence losses as $\mathcal{L}_{\text{KL}}$.
To instil discriminative knowledge, we train the invariant component with a triplet-loss ($\mathcal{L}_{\text{Tri}}^{z_{inv}}$) objective~\cite{yu2016sketch} where the distance of $z_{inv}$ extracted from a sketch (denoted as $s$), is reduced from that of its matching photo ($p+$), and increased from that of a non-matching one ($p-$). Furthermore, we resort towards discriminative sample generation by imposing a similar triplet objective on the synthesised embedding features $z_f$. We thus have:
\vspace{-0.2cm}
\begin{equation}
\label{equ: tripletz}
\begin{aligned} 
    &\mathcal{L}_{\text{Tri}}^{z_{inv}} =  \max \{0, m^{z_{inv}} + \delta(z_{inv}^{s}, z_{inv}^{p+}) - \delta(z_{inv}^{s}, z_{inv}^{p-})\},      
    \\&\mathcal{L}_{Tri}^{z_{f}} =  max \{0, m^{z_f} + \delta(z_{f}^{s}, z_{f}^{p+}) - \delta(z_{f}^{s}, z_{f}^{p-})\} 
\end{aligned}
\vspace{-0.0cm}
\end{equation}
where $m^{z_{inv}}, m^{z_f}$ are margin hyperparameters and $\delta(a,b) = ||a-b||^2$.
For simplicity, we have $\mathcal{L}_{\text{Tri}} = \mathcal{L}_{\text{Tri}}^{z_{inv}} + \mathcal{L}_{\text{Tri}}^{z_{f}}$.
Now the overall learning objective of our disentanglement VAE model is:
\vspace{-0.2cm}
\begin{equation}
\label{equ: omegavae}
    \mathcal{L}_\Omega = \mathcal{L}_{\text{rec}}
    + \lambda_1 \cdot \mathcal{L}_{\text{KL}} 
    + \lambda_2 \cdot \mathcal{L}_{\text{Tri}},
\vspace{-0.1cm}
\end{equation}
where $\Omega=\{\phi,\theta\}$ ; $\lambda_1, \lambda_2$ are weighting hyperparameters.

\subsection{Meta-Learning for Adaptive Disentanglement}
\noindent \textbf{Overview:} Disentangling styles in sketches is more challenging compared to other images like paintings~\cite{kotovenko2019content}. Besides having sparse visual cues, untrained amateur sketches hold much more variation in style unlike paintings which hold a distinct style-signature being trained under various definite schools of arts. More importantly, the exhibited style even for the same user can vary depending on which object instance depicted. It is thus critical to learn a disentanglement model that is capable of dynamically adapting to any unseen user style as well as object instances. This is achieved through meta-learning.

\noindent \textbf{Task Sampling:}  In a meta-learning framework \cite{tim2020metaSurvey}, a model is trained from various related labelled tasks. To sample a task $\mathcal{T}_{i} \sim p(\mathcal{T})$ here, we first select a random category $C_i$ out of $M$ categories. Out of all $n_i$ sketch-photo pairs in $C_i$, `$r_i$' randomly chosen pairs are set aside for validation (query) set ($\mathcal{D}^{val}_i$), while the remaining $N_i$ pairs constitute the training (query) set ($\mathcal{D}^{trn}_i$).
Inner loop update is performed over $\mathcal{D}^{trn}$ with an aim to minimise the loss in the outer loop over $\mathcal{D}^{val}$. Within every set, hard negatives are chosen from rest M-1 categories ensuring completely dissimilar instances. Next, to prepare the VAE described earlier for meta-learning, we introduce two new components. 

\noindent \textbf{Meta-enhancing feature encoder:}
\label{sec:FTlayer}
Inspired from \cite{tseng2020cross}, the first new component is a set of feature-transformation (FT) layers plugged into the Encoder ($Enc_\phi$), with an aim to dynamically 
\cut{adjust the encoding according to the input photo/sketch.}
{minimise the style-variance in sketches.}
\cut{Such layers would essentially minimise the style-variance in sketches over episodic training in a meta-learning paradigm.}
These FT layers are added after the batch-normalisation layers in $Enc_\phi$. For an intermediate feature map $\mathcal{F} \in \mathbb{R}^{h' \times w' \times c}$ where $h'$,$w'$ and $c$ are height, width and number of channels respectively, we sample the bias ($\omega$) and scaling ($\eta$) terms
as :
$
    \omega \sim \mathcal{N}(0, \mathrm{{SmoothReLU}}(\phi_\omega))
    \; ; \;
    \eta \sim \mathcal{N}(1, \mathrm{{SmoothReLU}}(\phi_\eta))
$
where $\phi_\gamma = \{\phi_\omega,\phi_\eta\} \in \mathbb{R}^{1 \times 1 \times c}$ are hyper-parameters that signify the standard deviation of Gaussian distributions for sampling affine transformation parameters.
The activation thus changes to: $\hat{\mathcal{F}} = \eta \times \mathcal{F} + \omega $.
As determining hyper-parameters $\phi_\gamma$ empirically for every layer across different sketch-styles would be costly, we optimise them via  episodic training -- commonly adopted in meta-learning. Each training episode consist an inner loop and an outer one.  In the inner loop, 
the model is updated with a training loss:
\vspace{-0.5cm}
\begin{equation}
    \begin{aligned}
    \Omega{'} \leftarrow \Omega
    - \alpha \nabla_{\Omega}\mathcal{L}_{trn}
    (\overbrace{\{\phi,\phi_{\gamma}\}}^{Enc}, \theta ;\; \mathcal{D}^{trn}),
    \end{aligned}
    \vspace{-0.2cm}
\end{equation}
where $\alpha$ is the inner-loop learning rate. Then in the outer loop, the layers are pulled out and a loss ($\mathcal{L}_{tst}$) is calculated on the validation set using the modified parameters $\Omega'$. 
As $\mathcal{L}_{tst}$ denotes the efficiency of feature-transformation layer, we update $\phi_\gamma$ in the outer-loop (with a learning rate $\beta$): 
\vspace{-0.2cm}
\begin{equation}
\label{equ: metaFT}
    \phi_\gamma \leftarrow \phi_\gamma - \beta\nabla_{\Omega'}\sum_{\mathcal{T}_i} \mathcal{L}_{tst} (\Omega' ; \; \mathcal{D}^{val}).
\vspace{-0.2cm}
\end{equation}
\noindent \textbf{Meta-regularising Disentanglement:} To adapt the necessary extent of disentanglement, as the second new component,  we introduce an episodic regularisation of the disentangled modal-invariant latent representation $z_{inv}$ across tasks \cite{balaji2018metareg}.
Here the regulariser is denoted as $Reg(\cdot)$ that applies $\ell_1$ norm regularization to each of the parameters $\Omega_{inv}$ of $z_{inv}$.
In each training episode, a regularisation loss is imposed over the parameters $\Omega_{inv}$ of $z_{inv}$, as
\vspace{-0.2cm}
\begin{equation}
    \mathcal{L}_{reg} = Reg_\psi(\Omega_{inv}) = \sum_h \psi^{(h)}|\Omega_{inv}^{(h)}|.
\vspace{-0.3cm}
\end{equation}
$ \mathcal{L}_{reg}$ is added to the task loss that contributes to the inner loop update of the model i.e $\Omega' \leftarrow \Omega $. 
In the outer loop,  the loss is calculated with updated parameters of $\Omega'_{inv}$ which therefore reflects the usefulness  of the current regulariser. Consequently, its parameter $\psi$ is updated by $\mathcal{L}_{tst}$ as 
\vspace{-0.1cm}
\begin{equation}
    \psi \leftarrow \psi - \beta\nabla_{\Omega'}\sum_{\mathcal{T}_i} \mathcal{L}_{tst} (\Omega' ; \; \mathcal{D}^{val}).
\vspace{-0.2cm}
\end{equation}
This weighted $\ell_1$ loss denotes a learnable weight control mechanism, which adaptively modulates the proportion of semantic knowledge to be retained in $\Omega_{inv}$ for efficient disentanglement of the invariant semantic. As the same regulariser is trained across varying tasks in a meta-training paradigm, it is learnt to generalise onto any unseen task characterised by a new style for object category/instance. 

\noindent \textbf{Meta-Optimisation:} 
We summarise the overall meta-optimisation objective here from all the learning objectives discussed so far. Following Eq.~\ref{equ: omegavae} the model parameters $\Omega$ are updated to $\Omega'$ in the inner loop with overall meta-training loss as:
\vspace{-0.3cm}
\begin{equation}
\label{equ: inner-update}
    \begin{aligned}
        \mathcal{L}_{trn} &= 
        \mathcal{L}_{rec}  + \lambda_1 \cdot \mathcal{L}_{KL}
        + \lambda_2 \cdot\mathcal{L}_{Tri} + \lambda_3 \cdot \mathcal{L}_{Reg} \;,       \\
        \Omega' &\leftarrow \Omega - \alpha \nabla_{\Omega}\mathcal{L}_{trn}
        (\underbrace{\{\phi,\phi_{\gamma}\}}_{Enc}, \theta, \psi; \; \mathcal{D}^{trn} ).
    \end{aligned}
    \vspace{-0.2cm}
\end{equation}
%
With updated model parameters, a validation loss is computed over validation set ($\mathcal{D}^{val}$).
Here the meta-learning pipeline is trained alongside regularisation and feature transformation losses to optimise a combined loss. The optimisation objective for the outer loop is thus formulated as: 
\vspace{-0.3cm}
\begin{equation}\label{equ: outer-update}
\begin{aligned}
&\mathcal{L}_{tst} = \mathcal{L}_{rec} + \lambda_1 \cdot \mathcal{L}_{KL} + \lambda_2 \cdot \mathcal{L}_{Tri} \\
&\operatorname*{argmin}_{\Omega, \psi, \phi_\gamma} \quad
   \mathcal{L}_{tst} (\Omega',  \psi, \phi_\gamma ;\; \mathcal{D}^{val} ).
\end{aligned}
\vspace{-0.2cm}
\end{equation}
As $\Omega'$  depends on $\Omega$, $\psi$ and $\phi_\gamma$ via inner-loop update (Eq.~\ref{equ: inner-update}), a higher order gradient needs to be calculated for outer loop optimisation.  Notably, the model updates by averaging gradient over meta-batch size of $\mathrm{B}$ sampled tasks.  

\vspace{-0.2cm}
\section{Experiments }\label{sec:experiments}
\vspace{-0.1cm}
\keypoint{Datasets:} 
For category-level SBIR, two datasets are used. Following \cite{liu2017deep, zhang2018generative}, the first dataset used is Sketchy \cite{sangkloy2016sketchy} (extended) which contains $75k$ sketches across $125$ categories with about $73k$ images \cite{liu2017deep} in total. For the second dataset, sketches are taken from the TU-Berlin Extension \cite{eitz2012humans} which contains 250 object categories with 80 free-hand sketches per category. We further use 204,489 extended natural photo images of the same 250 TU-Berlin categories provided by \cite{zhang2016sketchnet} to construct the photo part of the dataset.
For both datasets we split photos from each category as $70:10:20$ for meta-training ($N_i$), meta-validation ($r_i$) and retrieval evaluation respectively, with the sketches split into the three sets in the same proportion. Note that there is no overlapping between the three sets,  meaning that sketch-styles used in evaluation are not seen during training.
For FG-SBIR, two publicly available datasets, QMUL-Chair-V2 and QMUL-Shoe-V2 \cite{yu2016sketch, sain2020cross} are used. They contain 2000 (400) and 6730 (2000) sketches (photos) respectively. Out of the photos, we keep 275 (100) for retrieval evaluation and use the rest for training, with 1150 (200) as meta-train and 575 (100) for meta-validation from QMUL ShoeV2 (ChairV2) datasets respectively. As both contain multiple sketches per photo-instance (we choose those photos having at least 3 sketches while training), they suit well to our motivation of modelling the diversity in sketch-styles. 
The input images (sketch/photo) were resized to $256\times 256$ and $299 \times 299$ for SBIR and FG-SBIR respectively.

\keypoint{Implementation Details:} 
\label{implementation} 
We implement our model in PyTorch on a 12GB TitanX GPU. We use InceptionV3~\cite{szegedy2016rethinking} as our encoder network. The decoder architecture \cut{is taken from \cite{pang2019generalising}  which} consists of a series of stride-2 convolutions with BatchNorm-Relu activation applied to every convolutional layer except in the output which has $\operatorname{tanh}$ for activation. The feature extracted from the encoder is projected into three 64 dimensional vectors signifying $\mu$, $\operatorname{log} \sigma^2$ and $z_{inv}$. 
In practice, while training we first warm up our basic cross-modal framework ($\S \ref{sec:cmd}$) for 20 epochs, before inserting $Reg_\psi(\cdot)$ and FT-layers for meta-optimisation (Eq. \ref{equ: inner-update}, \ref{equ: outer-update}). We use Adam optimiser in both inner and outer loops with learning rates of $0.0005$ and $0.0001$ respectively. Hyperparameters  $\lambda_{1\rightarrow 3}$ (determined empirically) are set to 0.001, 1.0, 0.7 respectively while $\lambda_1$ is increased with linear scheduling to 1.8 for the last 75 of 200 epochs, for better training stability. We use a meta-batch size of 16 and set $\mu^{z_{inv}}$ and $\mu^{z_f}$ to 0.5 and 0.3 respectively (further details in supplementary). 

\keypoint{Evaluation:}
\label{sec: eval}
Category-level SBIR is evaluated similar to \cite{liu2017deep} using mean average precision(MAP) and precision at top-rank 200 (P@200) . For FG-SBIR we use top-q (acc@q) accuracy.
We also design an unconventional metric solely for qualitative comparison of modelling sketch diversity in FG-SBIR. Out of `$m$' photos with `$k$' sketches per photo ($p_i$), we define average retrieval rank $\mathcal{R}_{avg}$ = $\frac{1}{m}\sum_i \mathcal{R}_i$ where $\mathcal{R}_i$ is the rank of retrieving `$p_i$' against sketch `$s_i$'. $\forall p_i$, let rank variance $\mathcal{V}_i$ = $\operatorname{variance}(\mathcal{R}^i_1,\mathcal{R}^i_2, \dots, \mathcal{R}^i_k)$ where $\mathcal{R}^i_j|^k_{j=1}$ is the retrieval rank of $p_i$ against its $j^{th}$ sketch-style. Accordingly, average rank variance $\mathcal{V}_{avg}$ = $\frac{1}{m}\sum_i \mathcal{V}_i$. 
Lower the value of $\mathcal{R}_{avg}$ and $\mathcal{V}_{avg}$, higher is the {score} and {consistency} in retrieval accuracy against varying styles per photo, respectively.

\begin{table}
\centering
\setlength{\tabcolsep}{2pt}
\caption{\normalsize{Comparative results of our model against other methods on FG-SBIR (D $\rightarrow$ disentanglement baselines).}}
\label{tab:quantitative_fgsbir}
\footnotesize
\begin{tabular}{clcccc}
\hline
\multicolumn{2}{c}{\multirow{2}{*}{Methods}} & \multicolumn{2}{c}{Chair-V2} & \multicolumn{2}{c}{Shoe-V2}
\\ \cline{3-6} 
             &    & acc.@1 & acc.@10 & acc.@1 & acc.@10
\\ \hline  
\rule{0pt}{2ex} 
\multirow{4}{*}{\rotatebox[origin=c]{90}{SOTA}}
& Triplet-SN ~\cite{yu2016sketch}    & 47.65  & 84.24  & 28.71  & 71.56  \\
& Triplet-Attn ~\cite{song2017deep}  & 53.41  & 87.56  & 31.74  & 75.78  \\
& Triplet-RL ~\cite{bhunia2020sketch}     & 56.54  & 89.61  & 34.10  & 78.82  \\
& CC-Gen ~\cite{pang2019generalising}     & 54.21  & 88.23  & 33.80  & 77.86  \\
\hline
\multirow{2}{*}{\rotatebox[origin=c]{90}{D }}
& D-TVAE \cite{ishfaq2018tvae} & 49.37 & 81.63 & 27.62 & 70.32 \\
& D-DVML \cite{lin2018deep} & 52.78 & 85.24 & 32.07 & 76.23 \\
\hline
\rule{0pt}{2ex} 
\multirow{4}{*}{\rotatebox[origin=c]{90}{{Others}}}
& B-Basic-SN          & 49.58 & 85.41 & 29.45 & 72.83 \\
& B-SN-Group          & 50.35 & 88.28 & 30.14 & 75.62 \\
& B-Cross-Modal
\cite{spurr2018cross} & 52.24 & 86.58 & 31.18 & 73.51 \\
& B-Meta-SN           & 53.57 & 87.69 & 32.74 & 76.92 \\

\hline
& Proposed        & \bf 62.86 & \bf 91.14 &  \bf 36.47 & \bf  81.83 \\
\hline
\end{tabular}%
\vspace{-0.1cm}
\end{table}

\begin{table}
\centering
\setlength{\tabcolsep}{3.5pt}
\caption{\normalsize{Comparative results of our model against other methods on SBIR (D $\rightarrow$ disentanglement baselines).}}
\label{tab:quantitative_sbir}
\footnotesize
\begin{tabular}{clcccc}
\hline
\multicolumn{2}{c}{\multirow{2}{*}{Methods}} & \multicolumn{2}{c}{Sketchy (ext)} & \multicolumn{2}{c}{TU Berlin (ext)}
\\ \cline{3-6} 
             &    & mAP & P@200   & mAP & P@200
\\ \hline  
\rule{0pt}{3ex}
\multirow{2}{*}{\rotatebox[origin=l]{90}{SOTA}}
& DSH (64 bit) ~\cite{li2017deeper}  & 0.711 & 0.858   & 0.521 & 0.655\\
\rule{0pt}{2ex}
& GDH (64 bit) ~\cite{zhang2018generative}     & 0.810 & 0.894  &0.690 &0.728 \\
\hline
\multirow{2}{*}{\rotatebox[origin=c]{90}{D }}
& D-TVAE \cite{ishfaq2018tvae}  & 0.695 & 0.839 &0.507 & 0.643 \\
& D-DVML \cite{lin2018deep} & 0.785 & 0.891 & 0.648 & 0.693 \\
\hline
\rule{0pt}{2ex} 
\multirow{4}{*}{\rotatebox[origin=c]{90}{{Others}}}
& B-Basic-SN           & 0.715 & 0.861 & 0.531 & 0.659 \\
& B-SN-Group          & 0.738 & 0.872 & 0.572 & 0.661 \\
& B-Cross-Modal
\cite{spurr2018cross}  & 0.763 & 0.884 & 0.622 & 0.688 \\
& B-Meta-SN           & 0.824 & 0.897 & 0.674 & 0.715 \\
\hline 
& Proposed           & \bf 0.905  & \bf 0.927 & \bf 0.778 & \bf 0.795 \\
\hline
\end{tabular}%
\vspace{-0.5cm}
\end{table}
\begin{figure*}[!htb] 
    \centering
        \includegraphics[width=\linewidth]{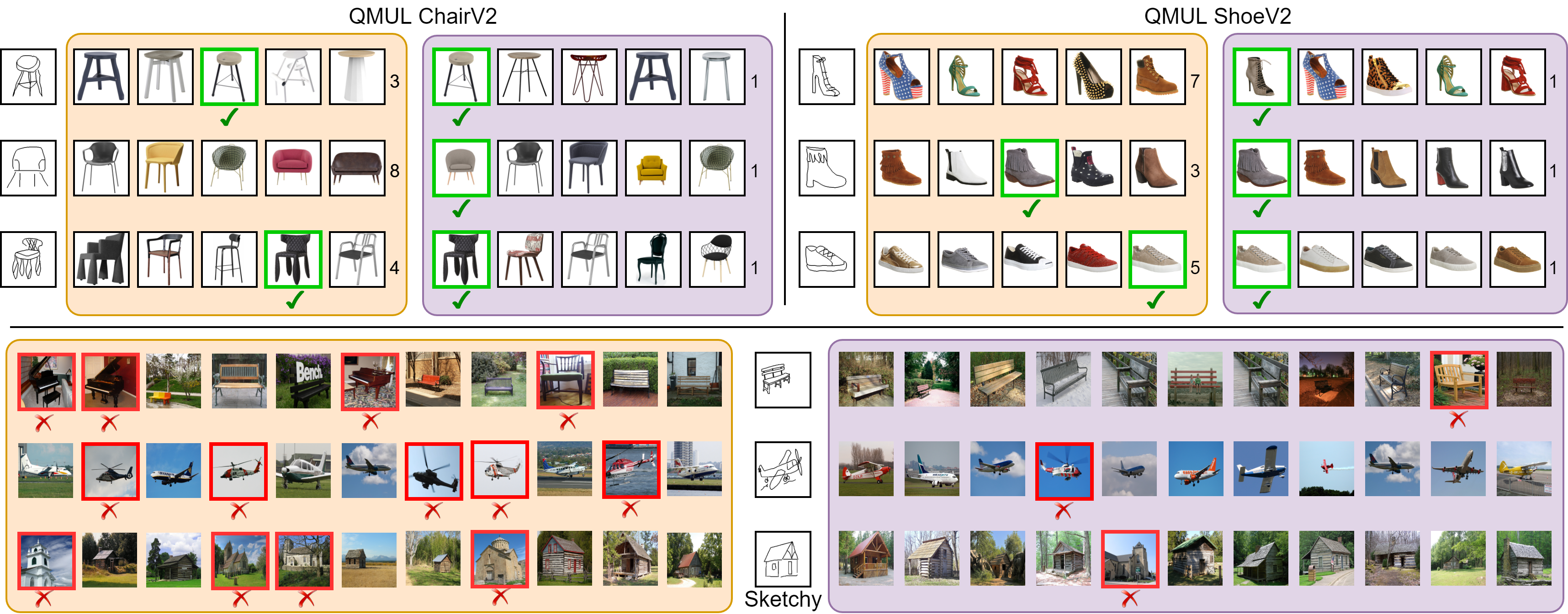}
        \caption{Qualitative retrieval results on QMUL ShoeV2, ChairV2 and Sketchy datasets. B-Basic-SN (orange) vs Ours (magenta). }
        \vspace{-0.4cm}
    \label{fig: mainRetrieval}
\end{figure*}
\vspace{-0.1cm}

\subsection{Competitors}
\vspace{-0.1cm}
\label{sec:competitors}
For both category-level and FG-SBIR, we evaluate our method against existing state-of-the-art (SOTA)  SBIR methods, and a few relevant latent representation disentanglement baselines adapted to our problem. These include:
\textbf{(a) SOTA:} 
\emph{Triplet-SN} \cite{yu2016sketch}  use Sketch-A-Net as baseline feature extractor trained using triplet loss. 
\emph{Triplet-Attn-SN} \cite{song2017deep} extended \cite{yu2016sketch} with spatial attention using a higher order HOLEF ranking loss.
\emph{CC-Gen}  \cite{pang2019generalising} takes a cross-category (CC) domain-generalisation approach, modelling a universal manifold of prototypical visual sketch traits that dynamically embeds sketch and photo, to generalise for unseen categories.
\emph{Triplet-RL} \cite{bhunia2020sketch} leverages triplet-loss based pre-training, followed by RL based fine-tuning for on-the-fly FG-SBIR. We report its results only on completed sketches as early retrieval is not our goal.
%
\emph{DSH} \cite{li2017deeper} unifies discrete binary code learning with visual sketch/photo feature maps to alleviate geometric distortion between sketches and photos.
\emph{GDH} \cite{zhang2018generative} learns a domain-migration network using binary hash codes in a generative adversarial paradigm with cycle-consistency losses, without relying on pixel-level alignment between cross-modal pairs.

\textbf{(b) Disentanglement methods}:
None of the existing SOTA SBIR models consider latent representation disentanglement. We therefore choose a number representative disentanglement methods and adapt them for  (FG-)SBIR for a fair comparison. For these models, encoded  features are used for distance-based retrieval during evaluation as done in our model.
\emph{D-TVAE} \cite{ishfaq2018tvae} uses a standard VAE training paradigm with single modality translation. A triplet loss is imposed on the extracted mean feature to bring the sketch and matching photo feature closer while distancing the negative photo-feature.
\emph{D-DVML} \cite{lin2018deep} employs a VAE framework with same-modality translation. It involves disentangling sketch features into invariant and variant components but the disentanglement operation is unregulated unlike ours. Besides VAE losses, the model is also guided by triplet loss objective between the invariant component of the sketch, its matching photo, and its non-matching photo.

\textbf{(c) Other relevant Baselines:} 
\emph{B-Basic-SN} is a naively built Siamese baseline similar to \textit{Triplet-SN} which replaces its Sketch-a-Net with Inception-V3 as a backbone feature extractor.
\emph{B-Cross-Modal} \cite{spurr2018cross} learns a cross-modal latent space in a VAE framework, involving translation amongst multiple (sketch and photo) modalities without disentanglement. We further impose a Triplet loss \cite{yu2016sketch} on the generated latent feature bringing sketch and matching photo features closer while distancing the negative one.
\emph{B-Group-SN} is similar to \textit{Triplet-SN} with Inception V3 \cite{sain2020cross} backbone feature extractor, where we concatenate feature embedding of three sketches against one corresponding category (SBIR) or photo-instance (FG-SBIR) and pass them through a linear layer to match the embedding dimension of the photo. This ensures that the sketch representation holds knowledge on multiple sketch-styles per object to some extent.
\emph{B-Meta-SN} simply employs vanilla MAML \cite{finn2017MAML} on a \textit{Triplet-SN} with an Inception-V3 backbone, without the FT layers or any disentangling regulariser.  It adapts using inner loop updates across retrieval tasks over categories in SBIR and over instances in FG-SBIR frameworks 
%
\subsection{Performance Analysis}
\label{sec:results}
\vspace{-0.15cm}
Comparative results on category-level and FG-SBIR are shown in  Table~\ref{tab:quantitative_sbir} and Table~\ref{tab:quantitative_fgsbir} respectively. The following observations can be made: \textbf{(i)} Our method outperforms all other compared methods under both settings and on all four datasets. This clearly illustrates the efficacy of the proposed method thanks to its ability of dynamically adapting to new user styles. \textbf{(ii)} The inferior results of \textit{Triplet-SN} and \textit{Triplet-Attn} are partially due to their apparently weaker backbone feature extractor of Sketch-A-Net. \textbf{(iii)} \textit{Triplet-RL} performs much better owing to its novel reward function designed in reinforcement setup towards sketch completion. \textit{CC-Gen} on the other hand comes close in performance, owing to its learning of universal manifold of visual traits aiding its generalising ability. However both \textit{Triplet-RL} and \textit{CC-Gen} ignore the style diversity issue. Compared to our model, their accuracy is lower by $6.32 (2.37)\%$ and $8.65 (2.67)\%$ for ChairV2 (ShoeV2) datasets respectively. \textbf{(iv)} For category-level SBIR, although both \textit{GDH} and \textit{DSH} perform well, they are clearly inferior to our method, as they rely on  only a singular sketch-image embedding via shared hashing network, without incorporating diverse sketch-styles belonging to the same object. \textbf{(v)} \textit{D-TVAE} performs poorly as it uses the same mean feature used for reconstruction as the modal-invariant component, thus offering sub-optimal disentanglement.
In contrast, \textit{D-DVML} fares better owing to better formulated guiding objectives and better modelling of the invariant component of a sketch/photo with higher discriminative knowledge instilled into the model. \textbf{(vi)} Being trained in a \textit{learning-to-learn} setup \textit{B-Meta-SN} performs better than its simpler counterpart \textit{B-Basic-SN} by $3.99$  $(3.29)\%$, but lags behind ours by $9.29$ ($3.73$)$\%$, as neither does it disentangle the stylistic variance, nor does it enforce style agnostic encoding of sketches. \textbf{(vii)} \textit{B-Cross-Modal} fares better than \textit{D-TVAE} as it harnesses greater information owing to learning a latent space aware of cross-modal knowledge. Without learning a disentangled feature space or style-agnostic encoding however, it performs poorer than our model.\cut{scores far lower than our model}
\textbf{(viii)} \textit{B-SN-Group} performs higher than \textit{Triplet-SN} with a boosted Acc@10 by $4.04 (4.06)\%$ as it now holds a stronger understanding of the search space with increased query knowledge. However, without disentanglement and meta-learning for better generalisation, it lags far behind our method.

\begin{figure}[!htb]
    \centering
        \includegraphics[width=\linewidth]{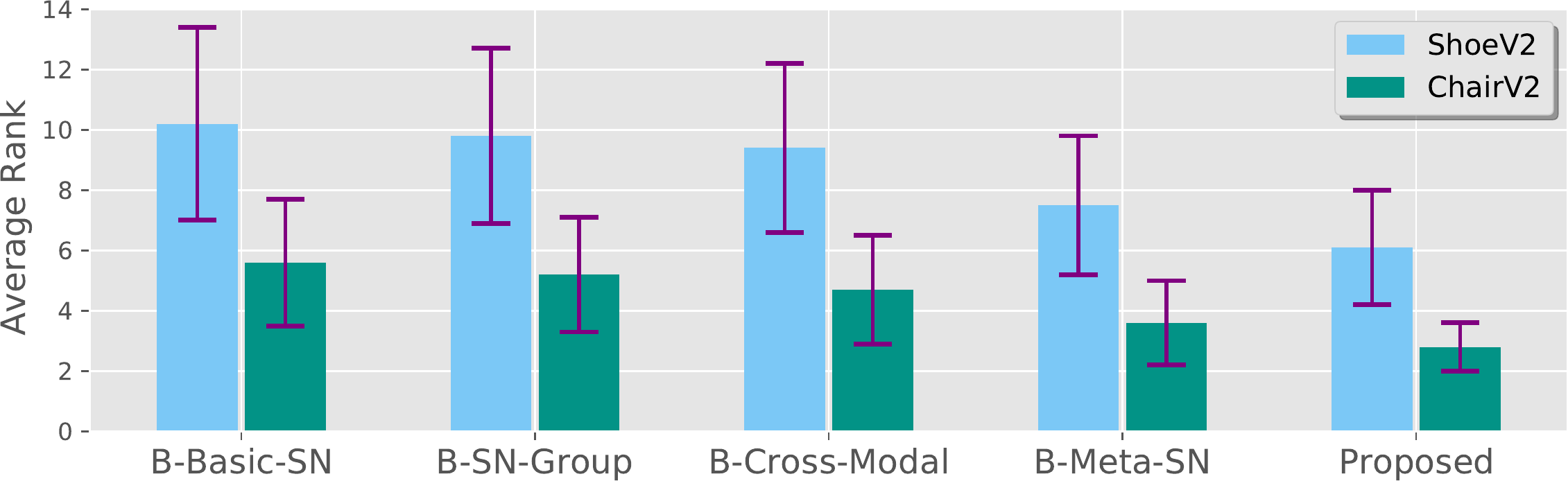}
      \caption{Figure shows proposed method to clearly surpass (lower is better) other baselines in both $\mathcal{R}_{avg}$ (bar height) and $\mathcal{V}_{avg}$ (variance line) in QMUL ShoeV2 and ChairV2 datasets.}
    \label{fig:plot}
     \vspace{-0.3cm}
\end{figure}

Diving deeper into the diversity modelling capability of our method, we plot the respective $\mathcal{R}_{avg}$ and $\mathcal{V}_{avg}$ values for some baselines (\textbf{B}) on QMUL ChairV2 and ShoeV2 datasets, obtained via our novel metric (\S\ref{sec: eval}-Evaluation) in Fig. \ref{fig:plot}. While the basic siamese net \textit{B-Basic-SN} shows a large variance among retrieval ranks of the same photo using its different sketches, our method has a lower rank variance in addition to a much lower average rank. This proves our method indeed models sketch-style diversity to a considerable extent, thus ensuring higher consistency in retrieval accuracy. Qualitative results are shown in Fig. \ref{fig: mainRetrieval}.
\vspace{-0.2cm}
\vspace{-0.1cm}
\subsection{Ablation Study}
\label{sec:ablation}
\begin{figure}[!htb]
\centering
\includegraphics[width=\linewidth]{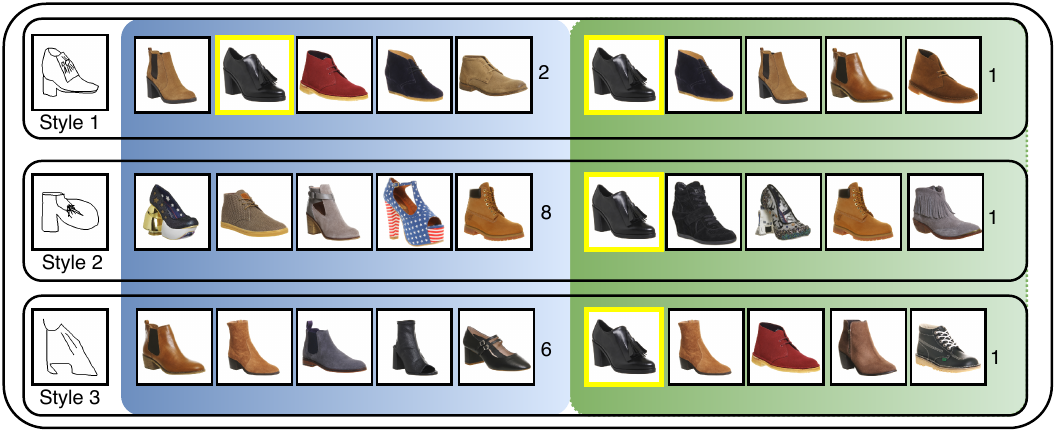}
\caption{Our (green) method's efficiency over \textit{w/o-Diversity} (blue) against different styles of the same shoe (highlighted in yellow) is shown. Numbers denote rank of the matching photo.}
\label{fig: diversity}
\vspace{-0.35cm}
\end{figure}

\vspace{-0.25cm}
\keypoint{Is modelling sketching diversity beneficial?}
For an in depth analysis we perform three experiments:
\textbf{(i)} A simple FG-SBIR baseline is trained similar to \textit{Triplet-SN}, on photos and two out of three styles of sketches per photo, keeping the third randomly chosen style per photo for evaluation. Here, the model is guided by triplet loss alone without disentanglement of the learned embedding space for dealing with style diversity. We show the results in Table \ref{tab:ablative} (\textbf{w/o Diversity}) on QMUL-ShoeV2 dataset~\cite{sain2020cross} as it has at least three sketches per photo. Results show that this model performs much poorer than our full model, showing the importance of style diversity modelling through explicit latent representation disentanglement. Some qualitative results with three different styles of the same object (unseen in training) are shown in Fig. \ref{fig: diversity} .
\textbf{(ii)} We hypothesise that true efficiency for our sketch representation can be judged in a sketch based sketch retrieval problem. Accordingly we train our model in same meta-learning paradigm in a single modality translation setup (no images) on QuickDraw dataset \cite{ha2017neural} with 345 categories, where we employ cross-style translation between sketches of same category, and triplet objectives similar to ours, with query sketch (s), its matching sketch (p+) and an unmatched (p-) one. The encoder with its modal-invariant semantic is used for retrieval. Following \cite{xu2018sketchmate}, we use 10k sketches for each category with 8:1:1 split for  meta-training, meta-test and evaluation respectively. An observed mAP (P@200) score of $\mathbf{0.748}$ $\mathbf{(0.792)}$ surpassing \cite{xu2018sketchmate} by $0.096$ $(0.103)$ shows credibility of our method.
\textbf{(iii)} Further validating the generality of our sketch-specific design, we perform sketch classification on Quickdraw dataset. We treat our style-agnsotic encoder as a feature extractor and fine tune it with a cascaded linear classification layer. A decent score of $\mathbf{75.81\%}$ against Sketch-A-Net's~\cite{yu2016sketchAnet} $68.71\%$, justifies the generality of our style-agnostic disentangled sketch representation.

\vspace{-0.25cm}
\keypoint{Importance of Feature-Transformation (FT) Layers:}
The reason for adding FT layers  to the VAE encoder is to impart style-agnostic encoding behavior to $Enc_\phi$. To validate this, we train our model without  the FT layers (\textbf{w/o FT} in Table \ref{tab:ablative}), and  observe a dip in performance by $3.19\%$ to $33.28\%$ on ShoeV2. Alternatively, one might empirically set the parameters of FT layers instead of meta-learning it. Accordingly we design a baseline where $\phi_\omega$ and $\phi_\eta$ are set to 0.6 and 0.25 respectively for all FT layers (\textbf{Fixed-FT}). Inferior results confirm the necessity of meta-learning the hyperparameters $\phi_\gamma$ instead of fixing them empirically. 

\noindent\textbf{Why Regularise Disentanglement?:}
Our regulariser $Reg_\psi(.)$ modifies the parameters of invariant feature predicting layer $z_{inv}$ to impart the knowledge regarding required extent of disentanglement for adaptively separating the variant and invariant components of sketches. To justify its significance we remove the regularising module and train the model keeping other modules intact (\textbf{w/o RegD} in Table \ref{tab:ablative}). We can see that the model's  Acc@1 drops to $32.57\%$ by $3.9\%$ on ShoeV2, which indicates that modelling the diversity in sketches is incomplete without dynamically adapting to the \textit{extent} of disentanglement. 

\begin{figure}[!hbt]
\begin{center}
  \includegraphics[width=\linewidth]{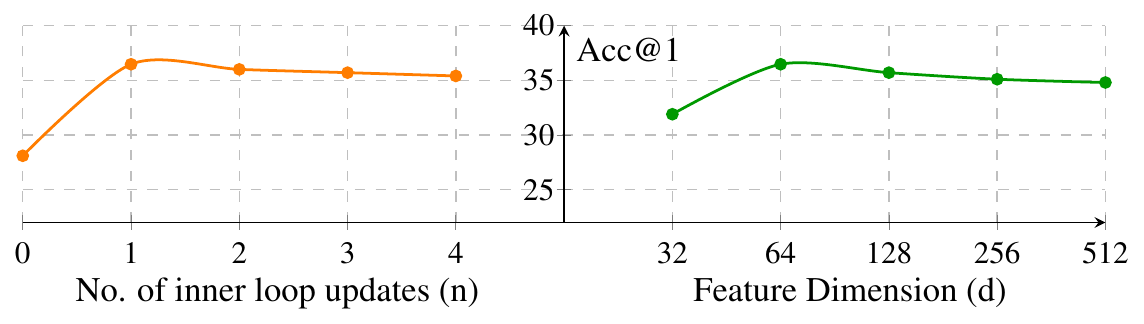}
\end{center}
\vspace{-0.5cm}
\caption{Varying (a) No. of Inner loop updates (optimal at n=1) (b) feature dimension (optimal at d=64) on QMUL Shoe-V2.}
\label{fig:ablation2}
\end{figure}

\noindent\textbf{Further Analysis:}\label{sec:further}
(i) Following \cite{finn2017MAML} we vary the number of inner loop updates (Eq. \ref{equ: inner-update}) during training. Fig. \ref{fig:ablation2} (left) shows a single step update to yield the best performance. Additional updates have a negative impact which might be due to inner loop over-fitting to unnecessary category-specific details of $\mathcal{D}^{trn}$, thus hampering the generic prior knowledge learned.
(ii) Evaluating our framework against varying encoded embedding space dimensions, we find optimal accuracy at d = 64, and that performance is stable with higher dimensions (Fig. \ref{fig:ablation2} right). 
(iii) Having similar evaluation setups, our time-cost (0.18/0.42 ms) for retrieval per query during evaluation on QMUL ChairV2/ShoeV2 lies close to that of \textit{B-Basic-SN} (0.14/0.37 ms).

\begin{table}
	\centering
 	\caption{Ablative Study}
 	\label{tab:ablative}
  	\resizebox{\columnwidth}{!}{
  	\footnotesize
	\begin{tabular}{lccccc}
	 
		\hline
        \multirow{2}{*}{Methods} & \multicolumn{2}{c}{Shoe-V2} & \multicolumn{2}{c}{Sketchy (ext)} \\ \cline{2-6} 
                & acc.@1 & acc.@10 & mAP & P@200  \\  
        \hline
        w/o Diversity   & 27.12 & 69.01  & | & | \\ 
        w/o MFT         & 33.28 & 75.34 & 0.852 & 0.916 \\ 
        w/o RegD        & 32.57 & 73.84 & 0.837 & 0.891 \\
        Fixed-FT        & 34.18 & 79.06 & 0.878 & 0.912 \\ 
        \hline
        Proposed        & \bf 36.47 & \bf  81.83 &  \bf 0.905  & \bf 0.927\\
		\hline
	\end{tabular}
 	} 
	\vspace{-0.5cm} 
\end{table}    
\vspace{-0.3cm} 

\vspace{-0.5cm} 
\section{Conclusion}
\vspace{-0.1cm} 
In this paper, we addressed  a key challenge for sketch-based image retrieval -- every person sketches the \textit{same} object \textit{differently}.  A novel style-agnostic SBIR model is proposed to explicitly account for the style diversity so that it can generalise onto unseen sketching styles. The model is based on a cross-modal VAE for disentangling a learned latent representation for photo/sketch into a modal-invariant part and a modal-specific part. To make such a disentanglement adaptive to unseen sketch styles, the model is meta-learned with two new components introduced for better generalisation. 
Extensive experiments show our method to outperform existing alternative approaches significantly.

{\small
\bibliographystyle{ieee_fullname}
\bibliography{egbib_main}
}
\cleardoublepage 
\onecolumn{
\centering
\LARGE
{Supplementary material for\\ StyleMeUp: Towards Style-Agnostic Sketch-Based Image Retrieval}  
\normalsize
\author{Aneeshan Sain\textsuperscript{1,2}  \hspace{.2cm} Ayan Kumar Bhunia\textsuperscript{1} \hspace{.2cm} Yongxin Yang\textsuperscript{1,2} \hspace{.05cm} \\ Tao Xiang\textsuperscript{1,2} \hspace{.2cm}  Yi-Zhe Song\textsuperscript{1,2} 
\\ \textsuperscript{1} SketchX, CVSSP, University of Surrey, United Kingdom \hspace{.08cm} \\
\textsuperscript{2} iFlyTek-Surrey Joint Research Centre
on Artificial Intelligence.\\
}}


\section*{Additional explanations}

\noindent \textbf{Clarity on bridging domain gap:}

From the viewpoint of bridging the domain gap, a gradient reversal layer is employed in Dey \etal \cite{dey2019doodle}, that is used to create a domain-agnostic embedding, which however does not differentiate if it comes from a sketch or a photo. Our motivation is different -- in addition to tackling the sketch-photo domain gap, we further focus on narrowing the domain gaps that exist amongst different sketching styles (i.e., learning a style-agnostic embedding). In particular, the \textit{feature transformation layer} helps bridge this style gap by simulating varying distributions in the intermediate layers of the encoder, and thus condition the encoder to \textit{generalise onto unseen sketching styles}. The meta-learning paradigm further ensures that this notion of style variance is minimised over episodic training, finally resulting in a style-agnostic embedding.\\

\noindent \textbf{Additional experimental comparison: }

The results of DSH and GDH on Sketchy and TU-Berlin have been taken directly from their respective papers. For further transparency we re-run these baselines using Inception-V3 as backbone. Table \ref{tab:my_label} shows these results to be in line with our conclusions for Sketchy and TUBerlin datasets respectively --

\begin{table}[hbt!]
    \centering
    \caption{\normalsize{Quantitative analysis using Inception-V3 backbone}}
    \vspace{0.1cm}
    \normalsize
    \begin{tabular}{ccccc}
         \hline
         \multirow{2}{*}{Method} & \multicolumn{2}{c}{Sketchy} & \multicolumn{2}{c}{TUBerlin}\\
         \cline{2-5}
         &        mAP  & P@200 & mAP   & P@200 \\
         \hline
         DSH   & 0.725 & 0.867 & 0.537 & 0.660\\
         GDH   & 0.821 & 0.896 & 0.696 & 0.741\\
         Ours  & \bf 0.905  & \bf 0.927 & \bf 0.778 & \bf 0.795 \\
         \hline
    \end{tabular}
    \label{tab:my_label}
\end{table}

\noindent \textbf{More on training details:}

The hyperparameters $\lambda_{1\rightarrow3}$ have been determined empirically. The impact of $\mathcal{L}_{KL}$ is suppressed ($\lambda_1$=$0.001$) during initial stages of training, and increased with linear scheduling later for better training stability. We further observed that $\lambda_2$ works best if kept constant throughout. Changing $\lambda_3$ had generally produced comparatively lower results. Margin hyperparameters for triplet losses $\mu^{z_{inv}}$ and $\mu^{z_{f}}$ were set empirically as well. 
Please note that unlike few-shot adaption in MAML, there is no adaptation step here during inference. Instead, meta-learning is employed only during training to learn a style-agnostic feature encoder for better generalisation.\\

\noindent \textbf{More on Fusing modal invariant and modal specific features:}

Combining these two components helps the model in keeping important details that might have been removed during disentanglement, for image (sketch/photo) reconstruction.  Furthermore, as we intend to learn \textit{how to disentangle} modal-invariant feature from modal-specific one, combining them to obtain a proper reconstruction re-verifies that the disentanglement itself has been learned properly. However, experimental results suggested that element-wise addition  performs better than concatenating the two components together. This is probably because the former establishes a clearer boundary between the disentangled components than concatenation.
\\


\end{document}